\pdfoutput=1
\documentclass[11pt]{article}

\usepackage[table]{xcolor}
\usepackage[preprint]{acl}

\usepackage{times}
\usepackage{latexsym}

\usepackage[T1]{fontenc}
\usepackage[utf8]{inputenc}

\usepackage{microtype}

\usepackage{inconsolata}

\usepackage{graphicx}

\usepackage{tabularx}
\usepackage{booktabs}
\usepackage{subcaption}
\usepackage{multirow}
\usepackage[capitalize]{cleveref}
\usepackage{enumitem}

\newcommand{\detect}{\textsc{Detect}}
\newcommand{\critique}{\textsc{Critique}}
\newcommand{\refine}{\textsc{Refine}}
\newcommand{\rerank}{\textsc{Rerank}}
\newcommand{\generate}{\textsc{Generate}}
\newcommand{\recipe}{\textsc{MAMM-Refine}}
\newcommand{\gptos}{2xG}
\newcommand{\claudes}{2xC}

\newcommand{\myparagraph}[1]{\vspace{0.3em}\noindent\textbf{#1}}

\title{\recipe{}: A Recipe for Improving Faithfulness in Generation with Multi-Agent Collaboration}

\author{David Wan \,\, Justin Chih-Yao Chen \,\, Elias Stengel-Eskin \,\, Mohit Bansal \\
UNC Chapel Hill\\
\texttt{\{davidwan,cychen,esteng,mbansal\}@cs.unc.edu} \\
}

\begin{document}
\maketitle
\begin{abstract}
Multi-agent collaboration among models has shown promise in reasoning tasks but is underexplored in long-form generation tasks like summarization and question-answering. We extend multi-agent multi-model reasoning to generation, specifically to improving faithfulness through refinement, i.e., revising model-generated outputs to remove factual inconsistencies. We investigate how iterative collaboration among multiple instances and types of large language models (LLMs) enhances subtasks in the refinement process, such as error detection, critiquing unfaithful sentences, and making corrections based on critiques. We design intrinsic evaluations for each subtask, with our findings indicating that both multi-agent (multiple instances) and multi-model (diverse LLM types) approaches benefit error detection and critiquing. Additionally, reframing critiquing and refinement as reranking rather than generation tasks improves multi-agent performance. We consolidate these insights into a final ``recipe'' called \textbf{M}ulti-\textbf{A}gent \textbf{M}ulti-\textbf{M}odel \textbf{Refine}ment (\recipe{}), where multi-agent and multi-model collaboration significantly boosts performance on three summarization datasets as well as on long-form question answering, demonstrating the effectiveness and generalizability of our recipe.\footnote{Our code is available at \url{https://github.com/meetdavidwan/mammrefine}.}
\end{abstract}

\section{Introduction}\label{sec:intro}
Large language models (LLMs) have achieved remarkable performance in natural language generation but still suffer from hallucinations and a lack of faithfulness \citep{guerreiro-etal-2023-hallucinations, zhang2023language, tang-etal-2023-understanding, tang-etal-2024-tofueval, liu-etal-2024-benchmarking}, where the generated content is inconsistent with the input source or the world. To address this problem, many studies have developed post-hoc self-refinement techniques \citep{NEURIPS2023_91edff07, gero2023selfverification, raunak-etal-2023-leveraging, jiang2023selfevolvecodeevolutionframework, gou2024critic, dcr}. However, these techniques have been found to be less effective without external feedback, as models require external information to identify errors \citep{huang2024large}. One promising avenue for extending models beyond their inherent capabilities is multi-agent debate \citep{chen2024reconcile, du2023debate, liang2023encouraging}, where multiple LLMs improve their answers over the course of a debate or discussion. The agents can be multiple instances of the same model or different models (i.e. multi-model). 

While several approaches focus on improving generation faithfulness through refinement, e.g. by breaking down the refinement process into fine-grained subtasks \cite{liu-etal-2023-improving, dcr}, past work has used a single instance of the same model for each of these subtasks, without multi-agent collaboration.  Different models, due to their diverse training data and methods, often exhibit different hallucination patterns \citep{rawte-etal-2023-troubling, guerreiro-etal-2023-hallucinations, ye2023cognitivemiragereviewhallucinations}. Therefore, adopting a multi-model, multi-agent framework could help systems achieve higher faithfulness by allowing models to revise their solutions based on diverse answers obtained through collaboration. In such collaborative settings, different models' hallucinations might cancel out.

However, several challenges remain before the promise of multi-agent and multi-model approaches can be realized in generation tasks: First, multi-agent frameworks have shown great promise for reasoning tasks \citep{chen2024reconcile, du2023debate, liang2023encouraging} where the final answers are generally from a closed set and easily verified, leading to easy stopping criteria and enabling voting across agents.  
Applying multi-agent collaboration to generative tasks such as summary refinement -- where final answers are long and difficult to verify -- is less straightforward. Additionally, due to the complexity and multitude of the design choices in generation and refinement tasks, it is not clear which components would benefit from a multi-agent framework. As verified by our empirical results, naively applying multi-agent reasoning to all subtasks might unnecessarily increase cost and could even hurt performance, as agents may lead each other down incorrect paths.

\begin{figure*}[!th]
    \centering
    \includegraphics[width=\linewidth]{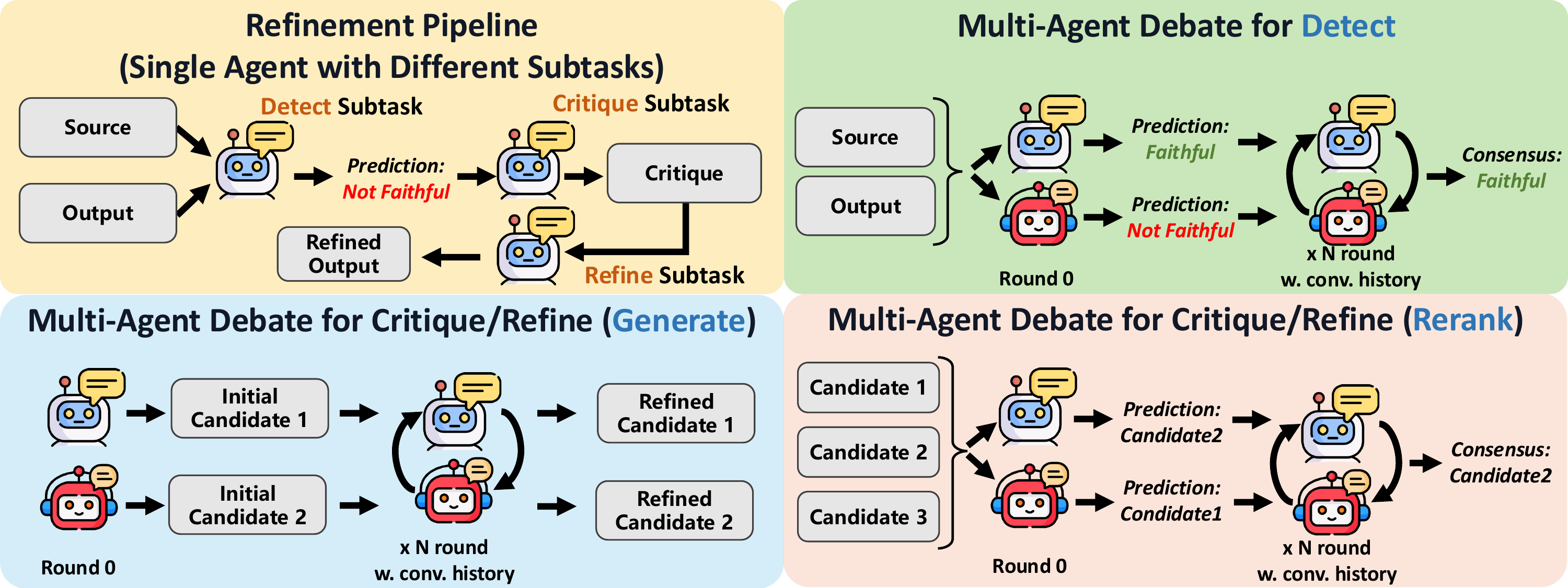}
    \caption{Illustration of the refinement pipeline (top-left) and how multi-agent debate is applied to different subtasks. In the \detect{} subtask (top-right), agents collectively choose among a discrete set of options, such as making yes/no decisions or selecting the most faithful candidate. For the \critique{} and \refine{} subtasks, we explore two approaches. In the bottom-left panel, we frame the task as generative (using \generate{}), where each agent updates its own critique or output based on other agents' responses. In the bottom-right panel, we frame it as a discriminative task using \rerank{}, where agents choose the best output from the candidates. While discriminative tasks converge to a single solution, generative tasks result in updated responses from each agent.}
    \label{fig:method_figure}
\end{figure*}

To extend multi-agent approaches to long-form generation, we focus on the task of improving faithfulness through refinement, as it is backed by extensive literature on evaluation metrics and generation strategies.
As illustrated in \autoref{fig:method_figure}, we conduct a comprehensive analysis to determine which refinement subtasks benefit most from a multi-agent pipeline. Focusing on the three-subtask approach from \citet{dcr}, a state-of-the-art refinement strategy, we consider \detect{}, \critique{}, and \refine{} subtasks, which can be recombined into different pipelines (e.g., \detect{}-\refine{}, \critique{}-\refine{}, \refine{} only). We apply multi-agent collaboration to these subtasks, framing \critique{} and \refine{} with two approaches: a discriminative method (\rerank{}) that selects the best option among multiple candidates, and \generate{}, which updates the answer freely.
Our research addresses three core questions:
\textbf{(1) Which refinement subtasks benefit from a multi-agent approach?} \textbf{(2) Which subtasks benefit from a multi-model approach?} \textbf{(3) For which task type (\generate{} or \rerank{}) is the multi-agent approach most effective?}

To answer these, we perform an extensive intrinsic analysis to find the optimal setting for each subtask, creating a ``recipe'', \textbf{M}ulti-\textbf{A}gent \textbf{M}ulti-\textbf{M}odel \textbf{Refine}ment (\recipe{}), that combines the best configurations. Using TofuEval \cite{tofueval}, a dataset with human-annotated sentence-level faithfulness judgments and critiques, we design intrinsic evaluation tasks for each of the three subtasks, \detect{}, \critique{}, and \refine{}, as illustrated in \autoref{fig:task_figure}.
Our findings show that multi-agent approaches generally outperform single-model baselines, with multi-model variants offering further gains, and for \critique{} and \refine{}, multi-agent methods provide consistent gains with \rerank{} but not with \generate{}.

Next, after determining the best configuration for each subtask from the intrinsic evaluations, we evaluate end-to-end performance on three summarization datasets with \recipe{}, MediaSum \cite{zhu-etal-2021-mediasum}, MeetingBank \cite{hu-etal-2023-meetingbank}, and UltraChat \cite{ding-etal-2023-enhancing},
comparing our approach with other refinement baselines. Across different refinement pipelines that use various combinations of subtasks, we show that selecting the best components from our intrinsic analysis gives us a generalizable recipe for improved refinement that holds across generation tasks and datasets, with gains on all three summarization tasks. We further show that our recipe generalizes to long-form question answering, improving faithfulness in a non-summarization domain.

\section{Method}\label{sec:method}
We begin refinement with a model-generated output $Y$ and optionally an input context $X$ (e.g., a summarization document or question-answering context \cite{xu-etal-2023-critical}). Using a refinement prompt and model $M_{r}$, we transform $Y$ into a refined output $Y_{r}$. We first outline common refinement pipelines and their corresponding subtasks, and next illustrate how each adapts to our multi-agent setting for generative tasks:

\myparagraph{Direct Refinement.}
A single task directly prompts the refinement model to improve the summary based on the document: $Y_{r} = M_{r}(X, Y)$.

\myparagraph{\detect{}-\critique{}-\refine{} (DCR).}
As illustrated in top left section of \autoref{fig:method_figure}, we follow \citet{dcr}'s breakdown of refinement into three steps, covering all components of the refinement process. First, each output sentence $y^{i}\in \{y^0, \ldots, y^N\}$  is evaluated by a detection subtask $M_{d}(X, y^{i})$, \detect{}, to produce a binary faithfulness label, indicating whether the sentence requires refinement. We define faithfulness as whether the output is supported by the input, and measure it using a model given the prompt described in Appendix~\ref{sec:prompts}.\footnote{Note that this prompting process is not the same what is used as the final evaluation metrics described in Section~\ref{sec:evaluation_tasks}.} In addition to a binary label, the detection step produces a reasoning chain which can be treated as a critique of the sentence (i.e. justifying why the sentence is unfaithful). For each sentence marked as unfaithful, we also optionally employ a critique subtask $M_{c}(X, y^{i})$, \critique{}, that generates a critique $c^{i}$ detailing the error span (i.e. which tokens make the output unfaithful) and suggest a fix. Finally, based on the outputs of \detect{} and \critique{}, we use \refine{} to generate a refined summary $Y_{r} = M_{r}(X, Y, C)$, where $C$ is the set of critiques, either directly from \detect{}'s reasoning or from the explicit \critique{} subtask. These three subtasks can be recombined into other variants, e.g. \detect{}-\refine{} (refining only on unfaithful generations), and \critique{}-\refine{} \cite{chern-etal-2023-improving} (refining based on critiques for all examples).

\myparagraph{Multi-agent Debate.} We adapt the multi-agent framework introduced by \citet{chen2024reconcile}, which has shown strong performance on short-form QA tasks such as commonsense and math reasoning. Let $A^{1}, \ldots, A^{n}$ be a list of $n$ agents participating in a discussion. In the initial round, we ask each agent to generate its own output $g^{i}_{0}$. For each subsequent round $k$, we ask each agent to update its answer based on all agents' responses from the last round, i.e., $g^{i}_{k} = A^{i}(g^{1}_{k-1}, \ldots, g^{n}_{k-1})$, forming a conversational state. That is, for each subtask, every agent can view the previous responses of the other agents and update its answer accordingly. Discussion ends when the maximum round is reached or when the agents have reached a consensus. 

\begin{figure*}[!th]
    \centering
    \includegraphics[width=\linewidth]{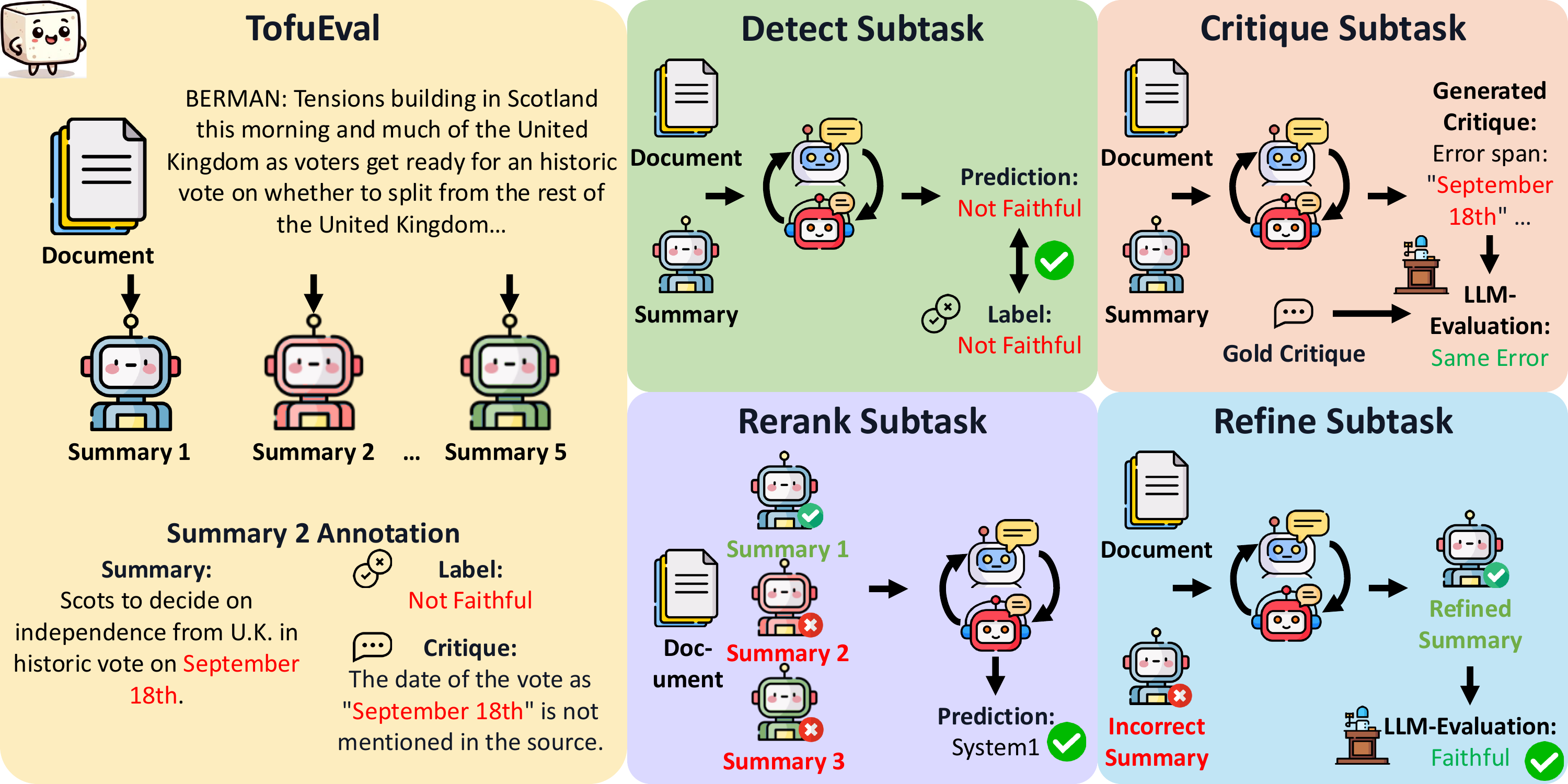}
    \caption{
    Illustration of our setup for intrinsic evaluations for different subtasks. 
    We convert TofuEval, a dataset of system summaries with human-annotated faithfulness labels and critiques, to tasks for evaluating the performance of different multi-agent setups for \detect{}, \critique{}, and \refine{} subtasks with \rerank{} and \generate{}.}
    \label{fig:task_figure}
\end{figure*}

\myparagraph{Extending to Generative Tasks.} 
While adapting multi-agent collaboration to a binary classification task like \detect{} is straightforward (see upper right of \autoref{fig:task_figure}), extending it to long-form tasks like \critique{} and \refine{} is challenging for two key reasons. First, evaluation is challenging as each agent produces its own answer; past work has addressed this by averaging the performance of individual agents \citep{du2023debate}. Secondly, determining a stopping criterion is challenging. Unlike with classification tasks, where it is clear when agents have converged to the same answer, evaluating and matching long-form outputs is a challenging open problem \citep{huang2024large}. Nevertheless, as shown in the bottom left of \autoref{fig:method_figure}, we experiment with a generative multi-agent variant (\generate{}) of \critique{} and \refine{}, where agents read others' answers and update their own. 

To better leverage the strength of multi-agent systems on closed-set tasks, we implement an alternative way to combine generations: \rerank{}, as illustrated in bottom right of \autoref{fig:method_figure}. Here, we transform open-ended generative tasks into a discriminative ones by asking agents to select the best generation from a set of candidates. Agents in \rerank{} produce item indices (a closed set), making the task a classification problem and simplifying voting and convergence checks.

\section{Experimental Setup}

\subsection{Agent Setup}
In all of our tasks, we use two strong agents of similar capability -- GPT-4o \cite{gpt4o} and Claude-3.5 Sonnet \cite{claude} when employing multiple agents; for our main experiments, we limit the number of agents to two to reduce computational cost. In Section~\ref{sec:gemini}, we also illustrate the gains achieved by adding more agents. We use the same prompts for all models, which are shown in Appendix~\ref{sec:prompts}. We consider the following combinations to evaluate the effectiveness of multi-agent settings along three axes: agent, model, and task. First, we differentiate between single-agent (SA) and multi-agent (MA) settings based on the number of agents used. Second, within the MA setting, we distinguish between single-model (SM), where multiple instances of the same model are used, and multi-model (MM), where different models serve as agents. From the pipeline perspective, we also consider a single-agent multi-model setting, where different subtasks use different models. Finally, we frame \critique{} and \refine{} as both generative (via \generate{}) and discriminative tasks (via \rerank{}). For the two models we employ, this results in two instances of GPT-4o (\gptos{}), two instances of Claude (\claudes{}), and the MAMM setting of using one GPT-4o and one Claude (G+C). For fair comparison between single-agent and multi-agent settings, we report the average performance of the single agents. For \generate{}, which generates multiple outputs, we report the average scores, similar to \citet{du2023debate}. For all tasks, we set the debate to run for a maximum of 10 rounds.

\subsection{Intrinsic Evaluation Tasks}\label{sec:evaluation_tasks}

We evaluate our methods using TofuEval \citep{tofueval}, a topic-focused summarization task with annotations on two datasets: MediaSum \cite{zhu-etal-2021-mediasum} and MeetingBank \cite{hu-etal-2023-meetingbank}. TofuEval contains 50 documents for each dataset, each paired with three topics. It also contains sentence-level faithfulness judgments from annotators for summaries generated by five different systems. For each sentence, annotators were asked to provide a binary faithfulness judgment and, if deemed unfaithful, write a critique explaining the error. We use this dataset to create intrinsic tasks for evaluating \detect{} and \critique{}. An example is shown on the left side of \autoref{fig:task_figure}, where summary 2 contains a hallucination regarding the date of the vote. For all four intrinsic evaluations, we use all 150 document-topic pairs (50 documents × 3 topics). We randomly selected one summary out of the five systems for each document-topic pair. We split the 50 documents into 10 for validation and 40 for testing, resulting in 30 validation and 120 test document-topic pairs. We tune all methods on the validation set. Our main research questions for the intrinsic evaluations are: (1) Does MA improve performance? (2) Does MM improve performance? (3) How do different frameworks affect performance? Appendix~\ref{sec:iteration_analysis} further explores how performance varies over debate iterations.

\myparagraph{\detect{} Evaluation.} We use the human-annotated faithfulness label and evaluate whether our detection model outputs the same label as a discriminative task. This yields 81 validation and 324 sentence-level test examples for MediaSum, and 85 validation and 328 test examples for MeetingBank. Following \citet{laban2021summacrevisitingnlibasedmodels}, we use balanced accuracy (BACC) to account for class imbalance. As baseline, we compare to strong automatic metrics, MiniCheck \cite{tang-etal-2024-minicheck} and AlignScore \cite{zha-etal-2023-alignscore}, which are trained entailment metrics between document chunks and summary sentences designed for summarization tasks. 

\myparagraph{\rerank{} Evaluation.} As an alternative to having agents directly debate their summaries, we aim to explore the best methods for reranking generated summaries. To achieve this, we use the human labels for all the different systems. Because TofuEval contains no gold summaries, we bootstrap this data by identifing cases where only one system's summary is judged by humans to be faithful, treating that summary as ``gold''. We then create test scenarios where we present this faithful summary along with two to four unfaithful summaries randomly sampled from the remaining summaries, resulting in sets of three to five summaries. We randomly shuffle the candidates to ensure the model is not biased toward any position. For evaluation, we measure the accuracy of the model in ranking the faithful summary highest from the set of candidates. We compare this to two baselines that use MiniCheck and AlignScore for reranking, selecting the output with the highest faithfulness score. 

\myparagraph{\critique{} Evaluation.} To evaluate the performance of the critique model, we consider two settings: \textit{Gold} and \textit{Detect}. These settings correspond to generating critiques when the summary sentence is considered unfaithful according to gold labels or model predictions, respectively. In the \textit{Gold} setting, we use the human-provided faithfulness labels, whereas in the \textit{Detect} setting, we use the predicted faithfulness labels from the best model found in the intrinsic \detect{} evaluation (G+C). We also evaluate the explanations generated by the \detect{} subtask as part of its chain-of-thought.
To evaluate our approach, we adopt the methodology outlined by \citet{dcr}, which we further verify with human evaluations in Section~\ref{sec:human_eval_critique}. Specifically, we prompt GPT-4o to assess whether the generated critique aligns with the human-written critique. We instruct the model to select one of the following options: (1) Error Match: The generated critique identifies the same error as described by the human. (2) Error, No Match: The generated critique discusses a different error than the one noted by the human. (3) No Error Detected, No Match: The generated critique states that there is no error, despite the human indicating otherwise.

\myparagraph{\refine{} Evaluation.}
We evaluate different methods for the refinement model using the same setup as in the final evaluation. We primarily test the effect of various refinement methods when using the best detector from the intrinsic \detect{} evaluation (G+C) and the best critique models for both the Gold critique (\claudes{}) and \detect{} settings (\gptos{}).
We assess the faithfulness of the summaries using MiniCheck \cite{tang-etal-2024-minicheck} and a GPT-4-based Likert evaluation, following \citet{dcr}. 
Both metrics show high correlations with human judgments of faithfulness \cite{tang-etal-2024-minicheck, liu-etal-2023-g, chiang-lee-2023-large, gao2023humanlikesummarizationevaluationchatgpt}, which we also verify in Section~\ref{sec:human_eval_metric}. 
We calculate the faithfulness of each summary sentence and then aggregate averaging across all sentences.

\subsection{Extrinsic Refinement Setup}
While the intrinsic tasks are tuned on the validation set of MediaSum and MeetingBank, we evaluate for the extrinsic evaluation on the test sets of MediaSum and MeetingBank, as well as on UltraChat \cite{ding-etal-2023-enhancing} as a held-out, out-of-domain setting. As baselines, we use each single agent individually to perform each component task. We then combine identical models in a multi-agent setting (e.g., \gptos{} or \claudes{}) and also explore a multi-model setting by combining Claude and GPT-4o. For tasks where a generative approach is applicable (i.e., critique and refinement), we further investigate \generate{}, as detailed in Section~\ref{sec:method}. We use MiniCheck and Likert scores for evaluation.

\begin{table*}[t!]
    \centering
    \small
    \begin{tabular}{ll cc | ccc ccc}
    \toprule
    && \multicolumn{2}{c|}{\detect{}} & \multicolumn{6}{c}{\rerank{}} \\
     & &  MediaSum & MeetingBank & \multicolumn{3}{c}{MediaSum} & \multicolumn{3}{c}{MeetingBank} \\
    Category & Method & & & 2 & 3 & 4 & 2 & 3 & 4 \\
    \midrule
    \multirow{2}{*}{Baseline} & MiniCheck & 72.8 & 69.8 & 56.7 &  46.7 & 53.3 & 80.0 & 53.3 & \textbf{76.7} \\
    & AlignScore & 70.8 & 71.6 & 56.7 & 46.7 & 46.7 & 70.0 & 63.3 &  70.0\\
    \midrule
    \multirow{2}{*}{Single Agent} & GPT-4o & 72.1 & 76.5 & 73.1 & 38.5 & 53.8 & 62.5 & 65.6 & 53.1 \\
    & Claude & 72.7 & 77.7 & 84.6 & 50.0 & 53.8 & 78.1 & 56.3 & 56.3  \\
    \midrule
    \multirow{2}{*}{Multi-Agent Single-Model} & \gptos{} & 72.5 & 75.5 & 83.3 & 46.2 & 40.0  & 62.5 & 66.7 & 50.0 \\
    & \claudes{} & 72.5 & 76.1 & \textbf{92.3} & 46.2 & \textbf{58.3} & \textbf{81.3} & 56.3 & 56.3\\
    \midrule
    \textbf{Multi-Agent Multi-Model} & \textbf{G+C}& \textbf{74.3} & \textbf{80.2} & \textbf{92.3} & \textbf{53.8} & 45.5 &\textbf{ 81.3} & \textbf{68.8} & 62.5 \\
    \bottomrule
    \end{tabular}
    \caption{ Detection (left) and reranking (right) results. We report balanced accuracy for detection and accuracy of selecting the faithful candidate for reranking (Acc@1). Reranking performance is broken down by number of distractors (columns). We \textbf{bold} the method that we select as the best method for \detect{} and \rerank{}.}
    \label{tab:discriminative_results}
\end{table*}

\section{Results}
We report the intrinsic and extrinsic results, and examine how \recipe{} generalizes to long-form question-answering. We provide additional discussions and show improvement from additional agents and how multi-agent performance changes after each round of discussion in \cref{sec:results_appendix}.

\subsection{\detect{} Intrinsic Results}
We present the best strategy for \detect{}, which identifies hallucinating sentences and thus helps the refinement systems to refine only where needed. We report BACC in the left side of \autoref{tab:discriminative_results}. We first note that the single agents perform competitively compared to the baseline of using the MiniCheck and AlignScore metrics to detect unfaithful sentences, especially on MeetingBank.

\myparagraph{Effect of Multi-Agent.} Using the same model does not improve performance, except for a slight improvement of $0.4\%$ on MediaSum when using \gptos{} over single GPT-4o. On MeetingBank, we observe a decrease of $1\%$ with both \gptos{} and Claude. 

\myparagraph{Effect of Multi-Model.}
Multi-model improves beyond the two base models. Specifically, G+C outperforms Claude, the best of the two single models, by $1.6\%$ and $2.5\%$ on MediaSum and MeetingBank, respectively. This indicates the effectiveness of multi-model in improving detection accuracy.

\myparagraph{Takeaway.} Multi-agent single-model does not improve \detect{}, but the multi-model variant helps.

\subsection{\rerank{} Intrinsic Results}
Next, we determine the best combination for \rerank{}, since it will be used for both critique and refinement. The accuracy of selecting the most faithful summaries with different numbers of candidates is shown on the right of \autoref{tab:discriminative_results}.

\myparagraph{Effect of Multi-Agent.}
Here, MA improves over its SA, specifically on MediaSum. However, we only observe an improvement of $3.2\%$ with re-ranking 2 candidates using \claudes{} on MeetingBank.

\myparagraph{Effect of Multi-Model.}
Similar to \detect{}, we find that MAMM generally achieves high accuracy: it is tied with MASM with 2xC for the highest accuracy when reranking two candidates, and outperforms all other variants when reranking three candidates on MediaSum and MeetingBank. This confirms the importance of having multiple models. Among the different settings, we find that the largest gain occurs when there are only two choices, improving accuracy by $7.7\%$ and $3.2\%$ on MediaSum and MeetingBank, respectively. This aligns with prior works showing that LLMs perform better in pairwise comparisons \cite{huang-etal-2024-embrace}.

\myparagraph{Takeaway.} The multi-model multi-agent approach improves reranking accuracy, showing the benefit of such a framework for closed-set tasks.

\subsection{\critique{} Intrinsic Results}
We present the results in \autoref{tab:critique_results}, which analyzes whether the generated critiques identify the same errors as the gold critiques. The critiques that come with the \detect{}'s CoT are overall worse than those from the dedicated critique subtask, where the highest error matching score with the two-step approach under the \textit{Detect} setting is $12\%$ higher. This underscores the importance of having an additional critique step, so as not to overload LLMs with two tasks at once \cite{dcr}.

\myparagraph{Effect of Multi-Agent.} For \textit{Gold} critique case, we observe that reranking on Claude's critiques performs the best, almost achieving a perfect score. This shows that \claudes{} can critique the correct problem if there is no error in \detect{}. Note that is unrealistic, as the \critique{} will not have perfect \detect{} predictions and thus will not have $0\%$ "No Error" outputs, where \critique{} fails to find errors. 
Using predictions from \detect{} gives us a more realistic idea of what a model will do when the sentence is actually correct and \critique{} incorrectly considers it having some faithfulness errors. Interestingly, for the more realistic \textit{Detect} scenario, reranking on \gptos{} critiques achieves the highest performance. Compared to a single GPT-4o setting, the multi-agent approach improves by $2.4\%$ in terms of capturing the correct error. Multi-agent approach performs the best under the two settings.

\myparagraph{Effect of Multi-Model.} For both \textit{Gold} and \textit{Detect} settings, G+C is ranked second. As it performs slightly worse than the \claudes{} for \textit{Gold} and \gptos{} for \textit{Detect}, MM still demonstrates its generalizability.

\myparagraph{Effect of Task Framing.} Finally, we also compare the generative task framing (\generate{}) and the discriminative framing (\rerank{}) in the bottom section of \autoref{tab:critique_results}. 
Overall, the best generative approach (\claudes{}) has a lower error matching rate than its reranking counterpart, which is the worst of the three multi-agent systems when reranked.

\myparagraph{Takeaway.} Though multi-model provides consistent improvement across the two settings, using single-model multi-agent to rerank critiques performs the best compared to other variants. \generate{} does not show improvement from multi-agent.

\begin{table}[!t]
    \centering
    \resizebox{\columnwidth}{!}{%
        \begin{tabular}{lll ccc}
        \toprule
        Setting & Category & $M_{C}$  & EM$\uparrow$ & EMM$\downarrow$ & NE$\downarrow$ \\
        \midrule
        \multirow{5}{*}{\begin{tabular}{ll}\textsc{Detect}'s\\CoT\end{tabular}} & \multirow{2}{*}{SA} & GPT-4o & 54.0 & 7.3 & 38.8 \\
        & & Claude & \underline{55.9} & 6.7 & 37.5\\
        \cmidrule{2-6}
        & \multirow{2}{*}{MASM} & \gptos{} & 51.9 & \textbf{8.7} & 39.6\\
        & & \claudes{} & 53.8 & \textbf{8.7} & 37.6\\
        & MAMM & G+C & \textbf{57.0} & \underline{8.9} & \textbf{34.1} \\
        \midrule
        \multirow{2}{*}{\begin{tabular}{l}\textit{Gold}\end{tabular}} & \multirow{2}{*}{SA} &  GPT-4o & 95.1 & 5.0 & 0.0 \\
         & & Claude & \underline{98.5} & \underline{1.6} & 0.0\\
         \cmidrule{2-6}
         \multirow{3}{*}{\begin{tabular}{ll}\textit{Gold}\\w. \rerank{}\end{tabular}} & MASM & \gptos{} & 96.8 & 3.3 & 0.0 \\
         & \textbf{MASM} & \textbf{\claudes{}} & \textbf{99.2} & \textbf{0.8} & 0.0 \\
         & MAMM &  G+C & 97.5 & 2.5 & 0.0 \\
         \midrule
        \multirow{2}{*}{\begin{tabular}{l}\textit{Detect}\end{tabular}} & \multirow{2}{*}{SA} & GPT-4o & 67.1 & 3.2 & 29.9 \\
        & & Claude & 68.3 & 2.5 & 29.3 \\
        \cmidrule{2-6}
        \multirow{3}{*}{\begin{tabular}{ll}\textit{Detect}\\w. \rerank{}\end{tabular}} & \textbf{MASM } & \textbf{\gptos{}} & \textbf{69.5} & \textbf{1.3} & \textbf{29.3} \\
        & MASM & \claudes{} & 68.3 & 2.5 & \textbf{29.3} \\
        & MAMM & G+C & \underline{68.9} & \underline{1.9} & \textbf{29.3}\\
        \cmidrule{2-6}
        \multirow{3}{*}{\begin{tabular}{ll}\textit{Detect}\\w. \generate{} \end{tabular}} & MASM & \gptos{} & 62.1 & 2.9 & 35.0 \\
        & MASM & \claudes{} & 67.5 & 0.9 & 31.6\\
        & MAMM & G+C & 66.1 & 2.0 & 31.9 \\
        \bottomrule
        \end{tabular}
    }
    \caption{\critique{} Results under \textit{Gold} and \textit{Detect} setting, and using \detect{}'s CoT. EM = Error Match, EMM = Error Mismatch, and NE=No Error Found. We \textbf{bold} the best strategy for \critique{} for the two settings.
    }
    \label{tab:critique_results}
\end{table}

\subsection{\refine{} Intrinsic Results}\label{sec:refine_intrinsic}
Next, we evaluate which method is best for refinement. We present the results of using \claudes{} critiques, as they achieve higher faithfulness scores with the validation set in Table~\ref{tab:refine_results} and report the results with \gptos{} critiques in Appendix~\ref{sec:refine_gpt4os}.\footnote{The result with \gptos{} critiques show the same trends as with \claudes{} critiques.}

\myparagraph{Effect of Multi-Agent.}
The best setting, \gptos{}, achieves only a $0.3\%$ gain in MiniCheck. We hypothesize that with good critiques, a strong LLM-based refinement model can perform the task well.

\myparagraph{Effect of Multi-Model.}
For G+C variant with \rerank{}, we similarly observe that it does not improve beyond the single-model performance. In fact, it achieves faithfulness scores between the two single-agent faithfulness scores.

\myparagraph{Effect of Task Framing.}
We also compare \rerank{} with \generate{} and find that the methods using \generate{} further hurt faithfulness when applied to \claudes{} and G+C, while providing a slight but not significant gain of $0.3\%$ over the method using \rerank{}.\footnote{We use paired bootstrap test \cite{koehn-2004-statistical}.} As mentioned in Section~\ref{sec:method}, \generate{} does not guarantee outputting a single candidate. Considering the limited improvement and the high computational cost of performing multiple rounds of \generate{} (since it requires generating outputs for all agents in each round) compared to only debating on the examples where agents choose different best candidates in \rerank{}, we opt for \gptos{} with \rerank{}.

\myparagraph{Takeaway.} Overall, we recommend refining using \gptos{} with \rerank{} on \claudes{} critiques, illustrating the need for multi-model approaches from the pipeline perspective, where different models excel at different tasks; that is, Claude excels at generating critiques, and GPT-4o excels at refinement.

\begin{table}[!t]
    \centering
    \small
    \begin{tabular}{cc |cc }
    \toprule
    Category & $M_{R}$ & MCS$\uparrow$ & GL$\uparrow$  \\
    \midrule
    Original & -  & 78.3 & 3.8 \\
    \midrule
    \multirow{2}{*}{Single Agent} & GPT-4o & 84.6 & 4.2 \\
    & Claude & 82.8 & 4.2 \\
    \midrule
    \textbf{MASM w. \rerank{}} & \textbf{\gptos{}} & \underline{84.9} & 4.2 \\
    MASM w. \rerank{} & \claudes{} & 82.5 & 4.2 \\
    MAMM w. \rerank{} & G+C & 83.4 & 4.2\\
    \midrule
    MASM w. \generate{} & \gptos{} & \textbf{85.2} & 4.2 \\
    MASM w. \generate{} & \claudes{} & 79.1 & 4.2 \\
    MAMM w. \generate{} & G+C & 81.4 & 4.2\\
    \bottomrule
    \end{tabular}
    \caption{\refine{} results with \claudes{} critiques with MiniCheck (MCS) and  GPT-4o Likert score (GL). We \textbf{bold} the method that we select as the best method for \refine{}. Full table with scores on MediaSum and MeetingBank separately is shown in \autoref{tab:refine_results_full}.
    }
    \label{tab:refine_results}
\end{table}

\begin{table*}[!t]
    \centering
    \small
        \begin{tabular}{lccc | cc cc cc }
        \toprule
        &&&& \multicolumn{2}{c}{MediaSum} & \multicolumn{2}{c}{MeetingBank} & \multicolumn{2}{c}{UltraChat} \\
         Method & $M_{D}$ & $M_{C}$ & $M_{R}$ & MCS$\uparrow$ & GL$\uparrow$  & MCS$\uparrow$ & GL$\uparrow$ & MCS$\uparrow$ & GL$\uparrow$  \\
        \midrule
        Original & - & - & - & $74.4^{\dagger}$ & $4.1^{\dagger}$ & $82.1^{\dagger}$ & $3.6^{\dagger}$ & 77.6 & $3.8^{\dagger}$ \\
        \midrule
        \multirow{2}{*}{\refine{} Only} & - & - & GPT-4o & 77.6 & 4.3 & $76.9^{\dagger}$ & $3.6^{\dagger}$ & 75.8 & $4.0^{\dagger}$ \\
         & - & - & \gptos{} & 78.3 & 4.3 & $77.9^{\dagger}$ & $3.7^{\dagger}$ & 76.7 & $4.1^{\dagger}$ \\
         \midrule
        \multirow{2}{*}{\detect{} + \refine{}} & Claude & - & GPT-4o & 77.4 & 4.3 & $81.9^{\dagger}$ & $3.7^{\dagger}$ & 78.3 & $4.0^{\dagger}$ \\
        & G+C & - & GPT-4o & 77.8 & 4.3 & $81.1^{\dagger}$ & $3.7^{\dagger}$ & 78.3 & $4.0^{\dagger}$ \\
        \midrule
        \multirow{3}{*}{\critique{} + \refine{}} & - & GPT-4o & GPT-4o & 78.9 & 4.5 & 85.1 & 3.9 & 80.0 & 4.3 \\
        & - & GPT-4o & \gptos{} & 78.9 & 4.5 & 85.2 & 4.0 & 80.6 & 4.3\\
        & - & \claudes{} & \gptos{} & 81.7 & 4.5 & 86.7 & 4.0 & 80.8 & 4.3 \\
        \midrule
        Single-Agent Single-Model & GPT-4o & GPT-4o & GPT-4o & 79.2 & 4.4 & 86.6 & \textbf{4.0} & 80.5 & $4.1^{\dagger}$ \\
        Single-Agent Multi-Model & Claude & Claude & GPT-4o & 78.7 & 4.4 & 86.1 & \textbf{4.0} & 81.3  & 4.2 \\
        Multi-Agent Single-Model & \gptos{} & \gptos{} & \gptos{} & 79.9 & 4.4 & 87.0 & \textbf{4.0} & 79.9 & 4.2 \\
        \recipe{} (Ours) & G+C & \claudes{} & \gptos{} & \textbf{82.4} & 4.4 & \textbf{87.4} & 3.9 & \textbf{81.5} & \textbf{4.3} \\
        \bottomrule
        \end{tabular}
     \caption{Results on MediaSum, MeetingBank and UltraChat with MiniCheck (MCS), GPT-4o Likert score (GL). We show the models used for \detect{} ($M_{D}$), \critique{} ($M_{C}$), and \refine{} ($M_{R}$). $\dagger$ denotes statistically significant improvement by \recipe{} over that entry ($p<0.05$ using paired bootstrap test).
     }
    \label{tab:results_main}
\end{table*}

\begin{table}[t]
    \centering
    \resizebox{\columnwidth}{!}{
        \begin{tabular}{cccc |cc}
        \toprule
        Method & $M_{D}$ & $M_{C}$ & $M_{R}$ & MCS$\uparrow$ & G-L$\uparrow$  \\
        \midrule
        Original & - & - & - & 76.7 & $3.5^{\dagger}$ \\
        SASM & G & G & G  &  80.1 & 3.9 \\
        SAMM & C & C & G  & 80.9 & 4.0  \\
        MASM & \gptos{} & \gptos{} & \gptos{} & 79.1 & 3.9 \\
        \recipe{} & G+C & \claudes{} & \gptos{} & \textbf{82.0} & \textbf{4.1}\\
        \bottomrule
        \end{tabular}
    }
    \caption{Results on Long-form QA with context.}
    \label{tab:lfqa}
\end{table}

\subsection{Overall Result with Final Recipe}
Finally, we evaluate \recipe{} on MediaSum, MeetingBank, and the held-out dataset, UltraChat. We first focus on applying our best configurations for each subtask to existing refinement pipelines. As shown in \citet{dcr}, direct refinement even degrades MiniCheck scores on MeetingBank and UltraChat, demonstrating the necessity of a pipeline with more fine-grained subtasks. 
Nevertheless, we also evaluate direct refinement and pipelines without all the fine-grained subtasks, showing that applying our best configuration of \refine{} and \detect{} subtasks improves the results, with results shown in \cref{tab:results_main}.
Specifically, using \gptos{} for $M_{R}$ improves MiniCheck by $1\%$ on MeetingBank and UltraChat, though still underperforming the original summaries, showing the need for critique-based refinement. Additionally, when we add \detect{}, our best MAMM setting (G+C) further improves over direct refinement. Similarly, for the variants of  \critique{}+\refine{}, switching to MA $M_{R}$ yields a slight gain, as observed in Section~\ref{sec:refine_intrinsic}. Specifically, \claudes{} for $M_{C}$ and \gptos{} for $M_{R}$ provides $2.8\%$, $0.8\%$, and $1.0\%$ boosts over using only GPT-4o for $M_{C}$ and $M_{R}$ on MiniCheck compared to the \critique{}+\refine{} baseline.

We finally examine the three-step approach using all of our best configurations. Here, we observe the highest MiniCheck scores. In fact, \recipe{} is the only method among three-step approaches that has a statistically significant ($p<0.05$) gain over the original summary on both MediaSum and MeetingBank, as measured by both metrics. On the UltraChat dataset, \recipe{} is also the only three-step variant with a statistically significant faithfulness improvement over the original summary according to the GPT-4o Likert score. We also test four settings -- applying single or multi-model configurations to single or multi-agent setups -- and evaluate these as an ablation study. For MediaSum and MeetingBank, multi-agent is important, while on UltraChat, multi-model is important. Nevertheless, we observe a consistent trend where both multi-agent and multi-model configurations are key to improving faithfulness.

\subsection{Extending to Long-form QA}\label{sec:lfqa}
Next, we also explore how the pipeline extends to other generation tasks, such as long-form question answering (LFQA). We use the ELI5 dataset \cite{fan2019eli5} collected in WebGPT \cite{DBLP:journals/corr/abs-2112-09332}, which includes questions, model-generated answers, and the corresponding supporting context. From this data, we randomly select 100 examples and apply our refinement model. With the supporting context, the task is essentially question-answering with retrieved evidence, i.e. retrieval-augmented generation. Since evaluating the faithfulness of LFQA with context has the same setup as summarization \cite{xu-etal-2023-critical}, we employ the same experimental setup and metrics. The results are shown in \autoref{tab:lfqa}. We observe that multi-model and multi-agent approaches improve the faithfulness of the answers, and our recipe provides the most faithful responses, improving $5.3\%$ on MiniCheck and $0.6$ points on the Likert score. We similarly observe as an ablation that multi-model provides a stronger gain than multi-agent. This illustrates that our recipe can not only generalize to a held-out summarization dataset, but to a held-out non-summarization generation task like long-form question answering. We also report the setup and results without the context in Appendix~\ref{sec:lfqa_appendix}.

\section{Related Work}
\myparagraph{Multi-Agent systems with LLMs.}
A large body of research focuses on multi-agent systems for reasoning tasks \citep{du2023debate, liang2023encouraging, yin2023exchange, chen2024reconcile, kim2024llmsproducefaithfulexplanations, haji2024improvingllmreasoningmultiagent, tang2024causalgptmultiagentapproachfaithful, sun2024detectingllmshallucinationmarkov}, where multiple LLMs engage in debates or discussions. Recent studies have also proposed multi-agent systems for LLM evaluation, where agents either undergo a peer review process, obtaining a win rate by ranking each other \citep{li2023prd}, or engage in debates to determine the better LLM response \citep{chan2023chateval}.
To address hallucination, \citet{feng-etal-2024-dont} propose using a multi-agent system to identify knowledge gaps between LLMs. The success of this paradigm hinges on the fact that reasoning tasks typically have well-defined solutions. In contrast, multi-agent systems for generation tasks largely focus on enhancing creativity through role-playing \citep{wang-etal-2024-unleashing, lu2024llm, li2023camel}, where evaluation metrics are less established. To the best of our knowledge, we are the first to propose a multi-agent long-form generation in the context of improving faithfulness on summarization and long-form question-answering.

\myparagraph{Refinement.}
Refinement has gained significant focus, including leveraging human feedback \cite{saunders-2023-improving} and automatic feedback through self-refinement from the same model \cite{NEURIPS2023_91edff07, gero2023selfverification, raunak-etal-2023-leveraging}, other trained models \cite{xu-etal-2024-llmrefine, akyurek-etal-2023-rl4f, paul-etal-2024-refiner, chern2023factoolfactualitydetectiongenerative, chen2024magicore}, or external tools \cite{jiang2023selfevolvecodeevolutionframework, olausson2024is, gou2024critic, chen2024teaching}. For improving faithfulness of summarization, many post-processing approaches \cite{fabbri-etal-2022-improving, balachandran-etal-2022-correcting, thorne-vlachos-2021-evidence} focus on training such refinement model, or using human-annotated numeric scores as feedback \cite{NEURIPS2020_1f89885d, wu2021recursivelysummarizingbookshuman, nguyen-etal-2022-make, scheurer2024traininglanguagemodelslanguage}. More recently, efforts have concentrated on using LLMs to directly refine generations, such as by utilizing fine-grained feedback from a faithfulness detector at the level of atomic, non-decomposable facts \cite{wan-etal-2024-acueval}, or employing a two-step \cite{liu-etal-2023-improving} or three-step \cite{dcr} refinement approach. Our work is complementary to past refinement and multi-LLM work, as we measure the effect of multi-agent approaches across the components of the refinement pipeline. 
By testing MM and MA settings, we create a generalizable refinement recipe across generation tasks.

\section{Conclusion}
We carefully curate components for incorporating multi-agent collaboration into generation, improving generation faithfulness through refinement. Through intrinsic evaluations, we find that employing multiple agents, particularly multiple models, benefits discriminative tasks like \detect{} and \rerank{}. We then show how to apply \rerank{} to \critique{} and \refine{}. In extrinsic evaluations, we find that the best variation for each component improves several refinement methods, and our final recipe shows gains on three summarization benchmarks and transfers to long-form question-answering tasks, showing its generalizability. 

\section*{Limitations}
First, our work primarily focuses on faithfulness, which is crucial to building user trust in LLMs and enabling safe model use. 
While there are other aspects, such as coherence and relevance, that could contribute to a comprehensive evaluation, we choose to evaluate faithfulness due to its rich literature and extensive experiments using the best automatic evaluation metrics. 
Regarding evaluation, although the automatic metrics we use have shown high correlations with human judgments of faithfulness, a gap still exists, which could be addressed by conducting human evaluations. 
However, considering the trade-off between the high cost and unreliability of using Mechanical Turk workers, we opt to report statistical significance based on automatic evaluations for more reliable assessments. Finally, refinement pipelines and multi-agent frameworks involve additional steps that lead to higher computational costs. However, these costs tend to reduce over time, and applying multi-agent reasoning to open-ended generation tasks more broadly is a crucial area for which we lay the groundwork. We do not forsee any particular risks beyond those inherent to any text generation task. Since our work focuses on improving faithfulness, it is aimed at mitigating some of the risks associated with using LLMs for generation. 

\section*{Acknowledgement}
We thank the anonymous reviewers for their helpful comments. This work was supported by Google PhD Fellowship, the Microsoft Accelerate Foundation Models Research (AFMR) grant program, an NSF-CAREER Award 1846185, DARPA ECOLE Program No. HR00112390060, and NSF-AI Engage Institute DRL-2112635. 
Any opinions, findings, and conclusions or recommendations in this work are those of the author(s) and do not necessarily reflect the views of the sponsors.

\bibliography{custom}

\appendix

\section{Experimental Setup Details}\label{sec:expermental_setup_appendix}
In our experiments, we first test multi-agent and multi-model approaches to each component separately using intrinsic evaluations, and then combine these components and measure end-to-end refinement performance. Here, we describe the setup of our intrinsic evaluations for the different subtasks, as shown in \autoref{fig:task_figure}, and then detail our final evaluation setup on three summarization benchmarks.

\subsection{Models}
We use the latest versions of GPT-4o and Claude as of October 12, 2024. The number of parameters for these models has not been disclosed. We use the default decoding parameters for all models. For sentence splitting for \detect{}, we utilize NLTK's library \cite{bird2009natural}.

\subsection{Datasets}
We use annotations from TofuEval on the MediaSum and MeetingBank, released under the MIT-0 license. UltraChat and WebGPT are released under the MIT license. We follow the authors' instructions to process the data. To our knowledge, the authors of the datasets ensured that there are no harmful data. All datasets are in English.

\subsection{Metrics}
We use MiniCheck and AlignScore, following the authors' original repositories. For GPT-4 Likert, we use the \textit{gpt-4-0125} version of GPT-4. For VeriScore, we use the authors' original code\footnote{\url{https://github.com/Yixiao-Song/VeriScore}} and employ GPT-4o for extracting and verifying claims.

\begin{table*}[!t]
    \centering
    \resizebox{\textwidth}{!}{
        \begin{tabular}{cc | cccccc | cccccc }
        \toprule
        & & \multicolumn{6}{c|}{\gptos{} Critiques} & \multicolumn{6}{c}{\claudes{} Critiques} \\
         & & \multicolumn{2}{c}{MediaSum} & \multicolumn{2}{c}{MeetingBank} & \multicolumn{2}{c|}{Average} & \multicolumn{2}{c}{MediaSum} & \multicolumn{2}{c}{MeetingBank} & \multicolumn{2}{c}{Average} \\ 
        Method & $M_{R}$ & MCS$\uparrow$ & G$\uparrow$  & MCS$\uparrow$ & G$\uparrow$ & MCS$\uparrow$ & G$\uparrow$ & MCS$\uparrow$ & G$\uparrow$  & MCS$\uparrow$ & G$\uparrow$ & MCS$\uparrow$ & G$\uparrow$  \\
        \midrule
        Original & - &  74.4 & 4.1 & 82.1 & 3.6 & 78.3 & 3.8 & 74.4 & 4.1 & 82.1 & 3.6 & 78.3 & 3.8 \\
        \midrule
        \multirow{2}{*}{Single Agent} & GPT-4o & 79.6 & 4.4 & 86.9 & 4.0 & \underline{83.2} & 4.2 & 82.1 & 4.4 & 87.0 & 3.9 & \underline{84.6} & 4.2 \\
        & Claude & 79.7 & 4.4 & 84.0 & 3.9 & 81.8 & 4.2 & 80.7 & 4.4 & 85.0 & 4.0 & 82.8 & 4.2\\
        \midrule
        MASM w. \rerank{} & \gptos{} & 79.9 & 4.4 & \textbf{87.0} & 4.0 & \textbf{83.5} & 4.2 & \underline{82.4} & 4.4 & \underline{87.4} & 3.9 & \underline{84.9} & 4.2 \\
        MASM w. \rerank{} & \claudes{} & \textbf{81.0} & 4.4 & 84.5 & 4.0 & 82.8 & 4.2 & 80.4 & 4.4 & 84.7 & 4.0 & 82.5 & 4.2 \\
        MAMM w. \rerank{} & G+C  & \underline{80.0} & 4.4 & 85.9 & 4.0 & 83.0 & 4.2 & 81.9 & 4.4 & 85.0 & 3.9 & 83.4 & 4.2 \\
        \midrule
        MASM w. \generate{} & \gptos{}  & 79.9 & 4.4 & \underline{86.5} & 4.0 & \underline{83.2} & 4.2 & \textbf{82.5} & 4.4 & \textbf{87.8} & 4.0 & \textbf{85.2} & 4.2 \\
        MASM w. \generate{} & \claudes{} & 76.9 & 4.3 & 80.2 & 3.9 & 78.5  & 4.1 & 76.7 & 4.4 & 81.6 & 4.0 & 79.1 & 4.2 \\
        MAMM w. \generate{} & G+C & 78.0  &  4.3 & 82.6 & 3.9 & 80.3 & 4.1 &  78.9 & 4.4 & 83.9 & 4.0 & 81.4 & 4.2 \\
        \bottomrule
        \end{tabular}
    }
    \caption{Full refine results with \gptos{} and \claudes{} critiques.}
    \label{tab:refine_results_full}
\end{table*}

\section{Results Details}\label{sec:results_appendix}

\subsection{Full \refine{} results}\label{sec:refine_gpt4os}
We report the full results with \gptos{} and \claudes{} critiques in \autoref{tab:refine_results_full}. With \gptos{} critiques, we also observe that \rerank{} improves performance over the single-agent baseline. Interestingly, in this setting, we do not observe any average improvement across both metrics using \generate{}. This highlights the effectiveness and reliability of \rerank{} compared to \generate{}.

\begin{figure*}[th!]
    \centering
    \begin{subfigure}[t]{0.49\textwidth}
        \centering
        \includegraphics[width=\textwidth]{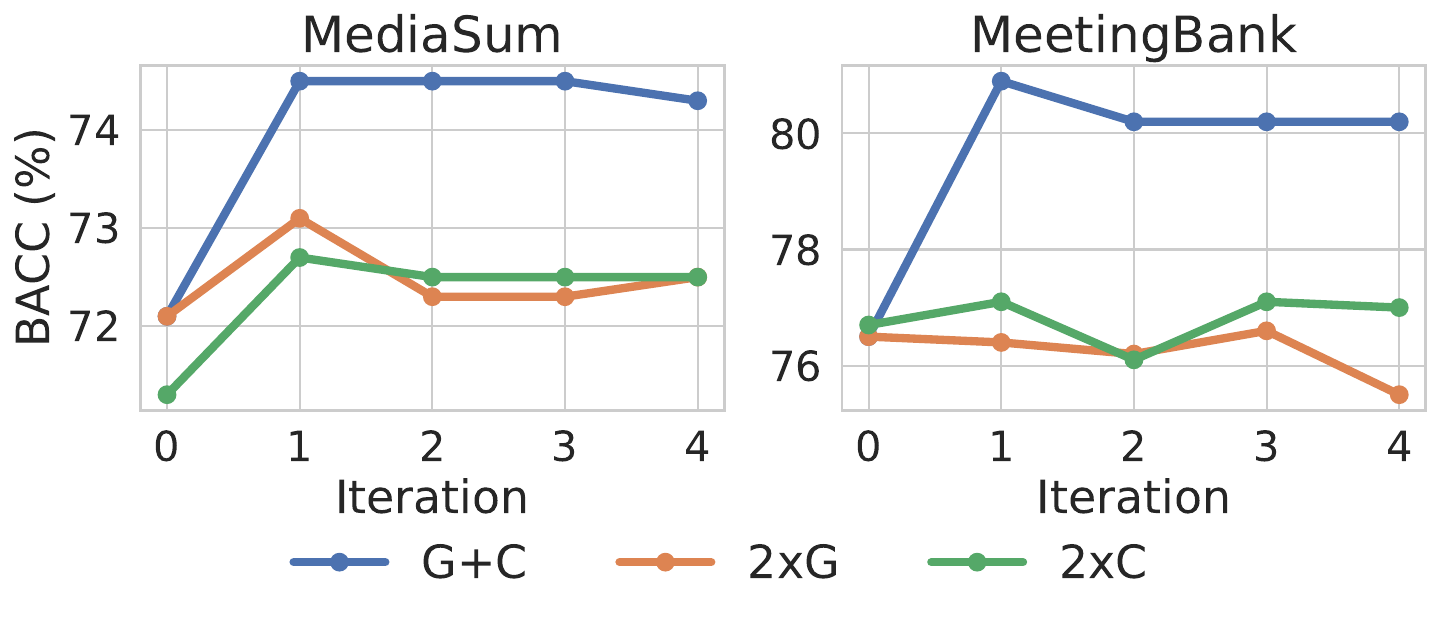}
        \caption{\detect{}}
    \end{subfigure}%
    \begin{subfigure}[t]{0.49\textwidth}
        \centering
        \includegraphics[width=\textwidth]{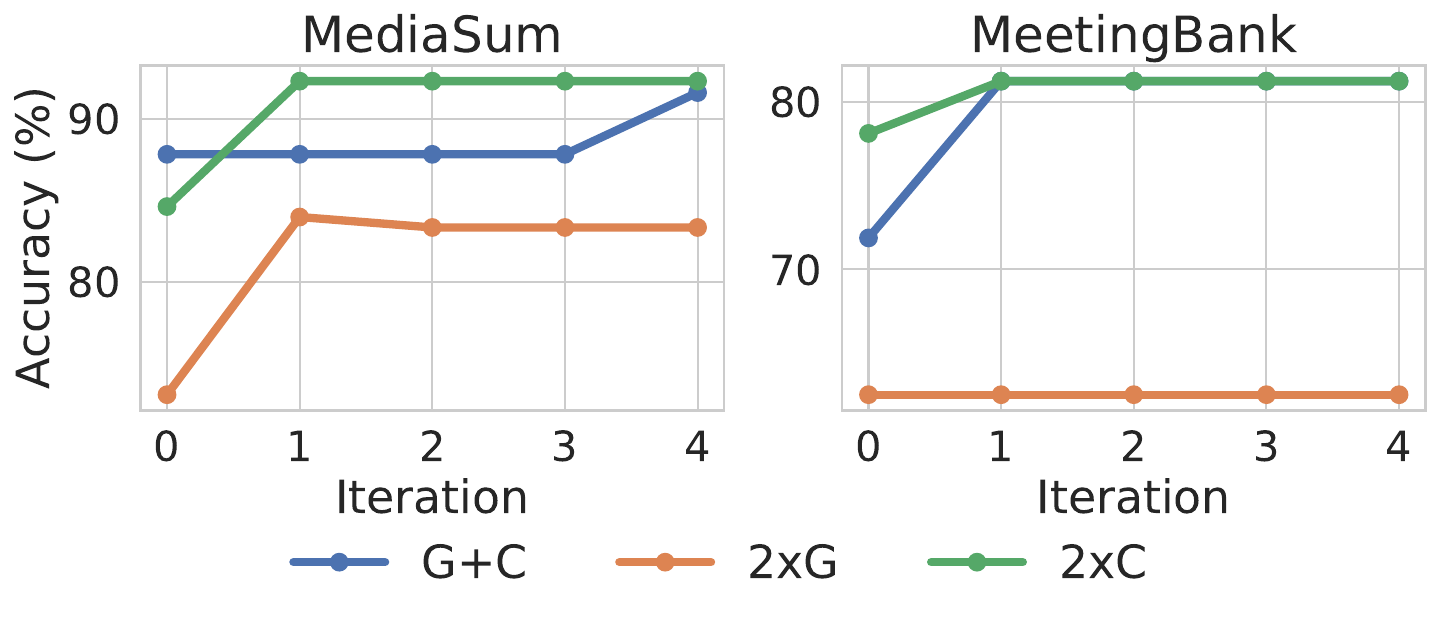}
        \caption{\rerank{}}
    \end{subfigure}%
    \caption{Detect and rerank multi-agent performance across multiple iterations.}
    \label{fig:discriminative_iterations}
\end{figure*}

\begin{figure*}[!t]
    \centering
    \begin{subfigure}[t]{0.49\textwidth}
        \centering
        \includegraphics[width=\textwidth]{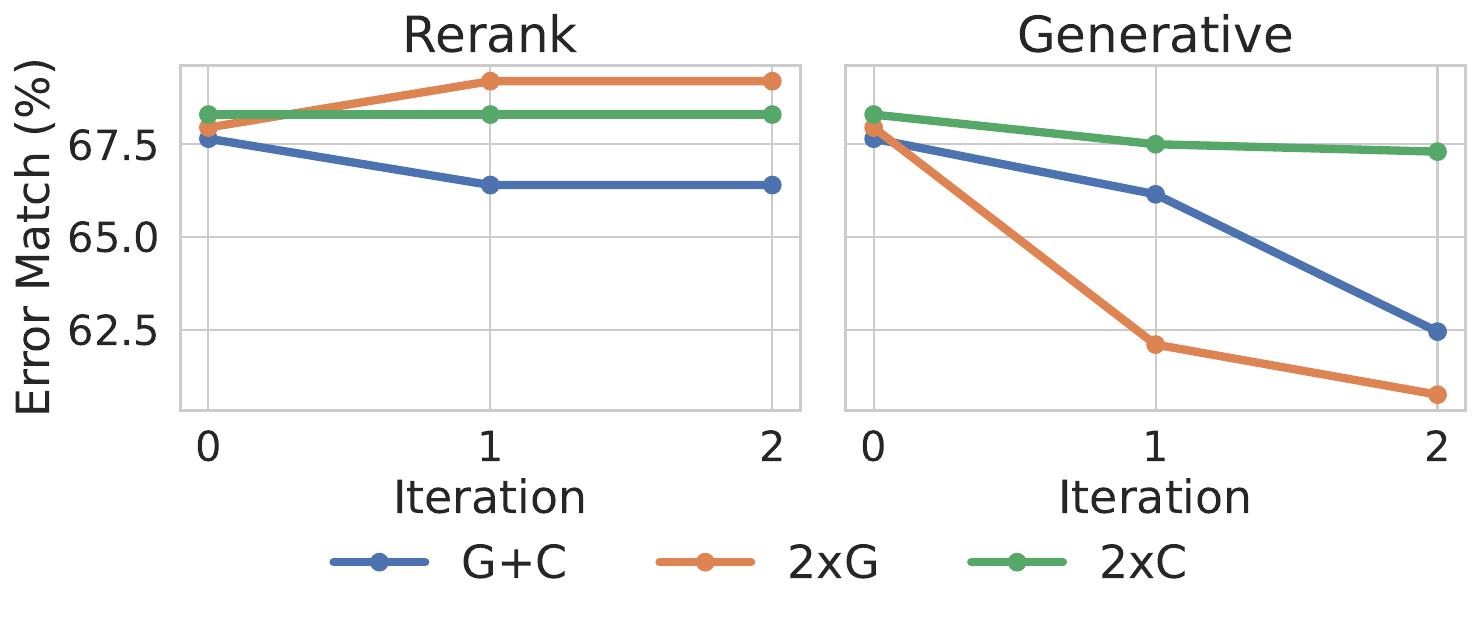}
        \caption{\critique{}}
    \end{subfigure}%
    \begin{subfigure}[t]{0.49\textwidth}
        \centering
        \includegraphics[width=\textwidth]{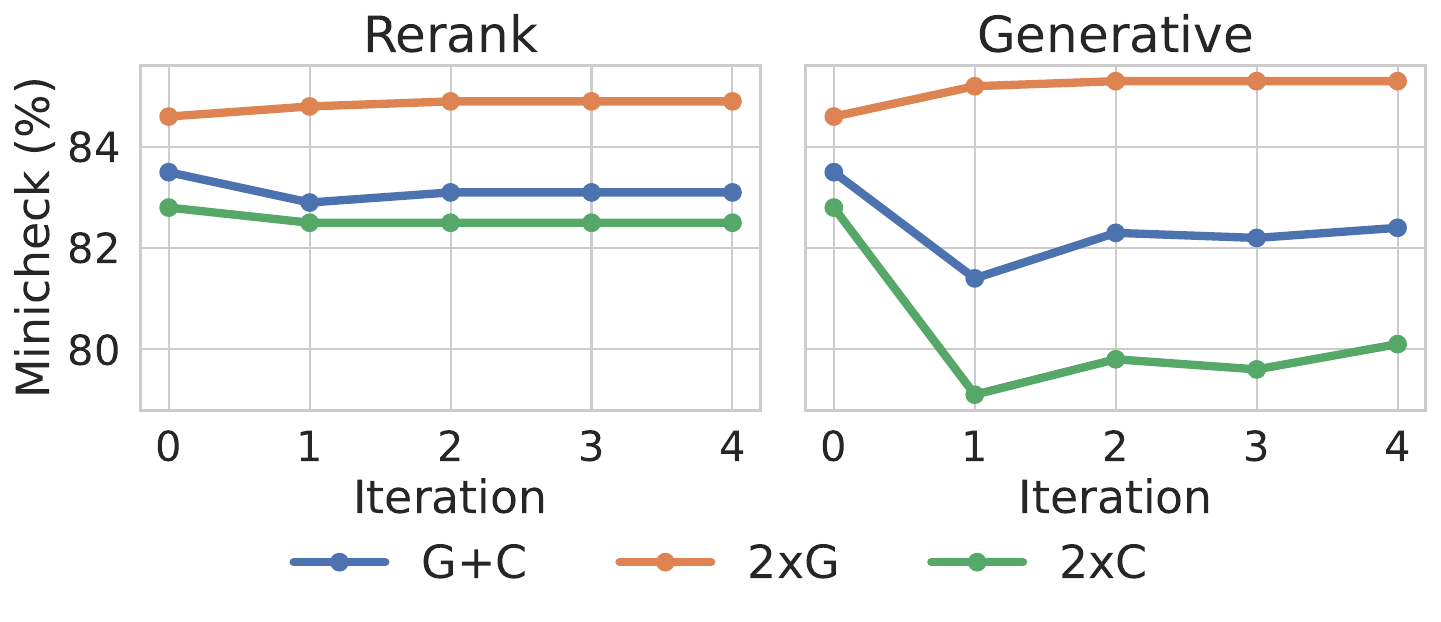}
        \caption{\refine{}}
    \end{subfigure}%
    \caption{Error match rate for \critique{} and faithfulness score for \refine{} across multiple iterations.}
    \label{fig:critique_iterations}
\end{figure*}

\subsection{Intrinsic Results on Multiple Iterations}\label{sec:iteration_analysis}

\myparagraph{\detect{}.} For the multi-agent models, we also analyze the balanced accuracy across multiple iterations, as the agents update their answers based on the other agent's response. 
As shown in \autoref{fig:discriminative_iterations}, one round of discussion provides the largest improvement, as the models show improvement in all cases except when using \gptos{} on MeetingBank. The largest improvement is with the multi-model setting, showcasing the benefit of having diverse responses. Additional rounds only address a few examples (e.g., 10/324 examples) and do not necessarily improve performance. 
We find that these are the harder examples on which the models have difficulty converging to an answer. See \cref{sec:detect_multi_round_analysis} for more details.

\myparagraph{\rerank{}.} Examining how accuracy changes over multiple rounds of debate, as shown on the right of \autoref{fig:discriminative_iterations}, we find that agents improve the most during the second round and converge by then, except for the multi-model multi-agent G+C method. In this case, on MediaSum, we observe additional improvement at round 3.

\myparagraph{\critique{}.} We further show the performance of the two frameworks across multiple iterations on the left of \autoref{fig:critique_iterations}. While reranking improves with further iterations, asking the models to continue refining their generations degrades performance.

\myparagraph{\refine{}.} When looking at the improvement across iterations on the right of Figure~\ref{fig:critique_iterations}, \gptos{} consistently improves slightly across multiple iterations. However, both multi-agent approaches (G+C and \claudes{}) performance decreases. With \generate{}, we observe a large decrease in faithfulness score, highlighting the more reliable performance of \rerank{}.

\begin{table}[!t]
    \centering
    \small
    \begin{tabular}{c c c}
    \toprule
    Round & \# Converged & BACC \\
    \midrule
    0 & 751 & 76.2 \\
    1 & 50 & 52.7 \\
    2 & 8 & 50.0 \\
    3 & 3 & 50.0 \\
    4 & 1 & 0.0 \\
    \bottomrule
    \end{tabular}
    \caption{Number of converged examples for each round and the corresponding BACC for the subset.}
    \label{tab:detect_multi_round_analysis}
\end{table}

\subsection{Analysis on Multi-Round for \detect{}}\label{sec:detect_multi_round_analysis}
To investigate whether the subsequent rounds involve harder examples that the agents have difficulty agreeing on, we calculate the balanced accuracy on the subset of examples, which the answer from the two agents finally converges for each round. We hypothesize that the model is unable to converge because both agents do not know the correct answer and thus the correct reasoning, making them incapable of convincing each other. We present the number of examples and the corresponding BACC for this subset in \autoref{tab:detect_multi_round_analysis}. We observe that for the 50 examples on which the agents converge in the first iteration, the BACC is already reduced from 76.2 to 52.7, and the remaining examples in the subsequent rounds only achieve accuracy at random chance levels. This indicates that the multi-agent approach can help improve performance on more examples but cannot improve cases where both agents are not confident.

\begin{table}[!t]
    \centering
    \small
    \begin{tabular}{ccc |cc }
    \toprule
    $M_{D}$ & $M_{C}$ & $M_{R}$ & MCS$\uparrow$ & GL$\uparrow$  \\
    \midrule
     G+C & \claudes{} & \gptos{}  & 84.9 & 4.2 \\
     \midrule
     G+C & G+C & G+C & 83.5 & 4.2 \\
     G+C+B & G+C+B & G+C+B & 84.2 & 4.3 \\
     G+C+B & \claudes{} & \gptos{} & 84.7 & 4.2 \\
    \bottomrule
    \end{tabular}
    \caption{Results with using three agents: GPT-4o (G), Claude (C), and Gemini (B). MCS=MiniCheck, GL=GPT4o 1-5 point Likert score.} 
    \label{tab:gemini_results}
\end{table}

\begin{table*}[!t]
    \centering
    \small
        \begin{tabular}{cccc | cc c}
        \toprule
        & & & & No Context & \multicolumn{2}{c}{With Context}\\
        Method & $M_{D}$ & $M_{C}$ & $M_{R}$ & VeriScore & MCS$\uparrow$ & G-Likert$\uparrow$  \\
        \midrule
        Original & - & - & - & 62.8 & 76.7 & $3.5^{\dagger}$ \\
        Single-Agent Single-Model & GPT-4o & GPT-4o & GPT-4o & \textbf{71.9} &  80.1 & 3.9 \\
        Single-Agent Multi-Model & Claude & Claude & GPT-4o & 71.4 & 80.9 & 4.0  \\
        Multi-Agent Single-Model & \gptos{} & \gptos{} & \gptos{} & 71.0 & 79.1 & 3.9 \\
        \recipe{} (Ours) & G+C & \claudes{} & \gptos{} & 70.2 & \textbf{82.0} & \textbf{4.1}\\
        \bottomrule
        \end{tabular}
    \caption{Results on Long-form QA for both with and without context.}
    \label{tab:lfqa_full}
\end{table*}

\subsection{Refinement with more Agents}\label{sec:gemini}
We also explore the use of three agents to assess the effect of increased agent diversity. We include Gemini-1.5-flash \cite{geminiteam2024gemini15unlockingmultimodal} as the third model, which performs similar to GPT4-o and Claude. Since the subtask that benefits the most from a multi-agent, multi-model approach is \detect{}, we experiment by adding the third agent to \detect{} only, as well as adding it to both \detect{} and \refine{} (and using the \rerank{} on the generations from the three agents). We present the results in \autoref{tab:gemini_results}. To measure the effect of adding a third agent for all subtasks, we compare the performance of using two agents across all subtasks with that of using three agents. We observe that using three agents provides additional gains in both the MiniCheck and Likert scores, indicating that having more models can indeed help. The three-agent version also achieves competitive scores with the variant where we use three agents for \detect{} and the best configuration from our recipe for \critique{} and \refine{}, indicating that having more agents may reduce the need for comprehensive testing to identify the best combination of subtasks.

The improvement observed with more agents can be attributed to more diverse outputs, each potentially containing different hallucinations due to the training paradigm. Alternatively, it can be thought of as the issue of hallucinations correlating with low confidence: Individual agents may produce hallucinations when they are less confident \cite{cao-etal-2022-hallucinated, van-der-poel-etal-2022-mutual}. However, the multi-agent framework mitigates hallucinations by enabling agents to collaborate and reach a consensus agreed upon by all (i.e., achieving high confidence), thereby improving faithfulness.

\subsection{LFQA Results without Context}\label{sec:lfqa_appendix}
For the case without context, the model must retrieve information from its own parametric knowledge. Here, we use VeriScore, a state-of-the-art verification model from \citet{song2024veriscoreevaluatingfactualityverifiable}. 

For the ``no context'' setting, which is reported in \autoref{tab:lfqa_full}, though single agent and single model performs the best, our recipe improves over the original answers by $6.3\%$. Among the different variations, single-agent multi-model outperforms multi-agent single-model, indicating that there is still benefit of using multiple models.

\begin{table}[!t]
    \centering
    \resizebox{\columnwidth}{!}{%
        \begin{tabular}{c c c c c}
        \toprule
        Method & Detect & \multicolumn{3}{c}{Critique} \\
        & BACC & EM$\uparrow$ & EMM$\downarrow$ & NE$\downarrow$ \\
        \midrule
         \rowcolor{gray!25}
        GPT-4o & 72.1 & 95.1 & 5.0 & 0.0 \\
        Llama3.1-8B & 62.2 & 67.2 & 32.8 & 0.0 \\
        \midrule
        MAMM w. Llama3.1-8B & 71.1 & 93.8 & 6.25 & 0.0 \\
        \bottomrule
        \end{tabular}
    }
    \caption{MediasSum results with using smaller model, Llama3.1-8b on \detect{} and \critique{}.}
    \label{tab:llama8b_results}
\end{table}

\subsection{Additional Analysis}
We have observed that the initial performance of each agent before the debate is crucial, as the debate outcomes are heavily influenced by these starting points. Specifically, when there is a large discrepancy between the agents' performances, combining them can help improve the weaker agent while maintaining similar (or slightly worse) performance for the better agent. This dynamic explains why certain settings work better for specific subtasks. For instance, in the critique task, GPT-4o and Claude single agents differ by 3.4 EM points, and MAMM averages their performances. However, SMMA effectively enhances the performance of both agents, and since Claude performs better initially, the MASM achieves the highest overall score by leveraging its stronger baseline.

In contrast, the refine task presents a scenario where GPT-4o outperforms Claude as a single agent. Here, SMMA benefits more from GPT-4o’s higher baseline, allowing it to refine and further improve its responses. Meanwhile, MAMM struggles due to Claude’s relatively lower performance, which drags down the combined results. These observations demonstrate that SMMA is better suited for tasks where one agent consistently outperforms the other, as it capitalizes on the stronger model's ability to refine its outputs during debate.

Qualitatively, we also find that generations from the same models exhibit little variation, so MA of the same model does not significantly aid in providing options for reranking. Additionally, \rerank{} can make mistakes when faced with two choices of differing quality, especially in cases where the topics are marginal in TofuEval - when dealing with less frequently mentioned information.

To verify this, we also test on MediaSum by running both larger and smaller models, where the performance discrepancy is more pronounced. Specifically, we employ the Llama3.1-8B model and GPT-4o as two agents and evaluate them using \detect{} and the gold setting of \critique{}. The results are shown in \autoref{tab:llama8b_results}. We observe that the smaller 8B model significantly underperforms compared to GPT-4o. When combined in the MAMM setting, the overall performance is slightly lower than that of GPT-4o alone. In cases where the two models disagree, the smaller model agrees with the larger model about $82\%$ of the time, thus failing to contribute to performance improvements. In the remaining $18\%$ of cases, the larger model is persuaded to accept the incorrect answer from the smaller model. These findings underscore the importance of using agents with similar performance levels to achieve further improvements on subtasks.

\begin{table*}[!t]
    \centering
    \small
    \begin{tabularx}{\textwidth}{p{1.5cm} X}
        \toprule
        Method & Prompt \\
        \midrule
        Direct \newline Refinement & I summarized the following document on the topic `\{Topic\}': \newline \{Document\}  \newline Summary of the above document on topic `\{Topic\}': \newline \{Summary\} \newline If there are any factual inconsistencies in the summary then edit the summary such that the refinement doesn't have any inconsistencies. Consistency in this context implies that all information presented in the summary is substantiated by the document. If the summary is consistent, then just the copy the same summary with no changes. When refining, make the minimum number of changes.\\
        \midrule
        \detect{} & Document: \newline \{Document\}\newline Sentence:\newline \{Sentence\} \newline Determine if the sentence is factually consistent with the document provided above. A sentence is factually consistent if it can be entailed (either stated or implied) by the document. Please briefly explain the reason within 50 words. Output your answer in json format, with the format as follows: \{\{``reasoning'': ``'', ``answer'': ``''\}\}. Please strictly output in JSON format. Only answer yes or no in the ``answer'' field.\\
        \midrule
        \rerank{} & Document: \newline \{Document\} \newline Summarize the provided document focusing on "{Topic}". The summary should be less than 50 words in length. \newline \#\#\# Summary 1: \{Summary1\} \newline \#\#\# Summary 2: \{Summary2\} \newline ... \newline Select the best summary that contains the least amount of factual inconsistencies. Consistency in this context implies that all information presented in the summary is substantiated by the document. Please briefly explain the reason within 50 words. Output your answer in json format, with the format as follows: \{\{``reasoning'': ``'', ``answer'': ``''\}\}. Please strictly output in JSON format. Only answer numbers in the "answer" field.\\
        \midrule
        \critique{} & I summarized the following document on the topic: `\{Topic\}': \newline \{Document\} \newline Summary of the above document on topic `\{Topic\}': \newline \{Summary\} \newline Reason about the factually inconsistent span in the sentence. A span is factually inconsistent if it cannot be substantiated by the document. Give reasons for the factual inconsistency, point to the error span by stating ``The error span: $\langle$span from sentence$\rangle$'' and end your answer with a suggested fix to the summary.\\
        \midrule
        \refine{} & I summarized the following document on the topic `\{Topic\}': \newline \{Document\} \newline Summary of the above document on topic `\{Topic\}': \newline \{Summary\} \newline Feedback for the above summary: \newline \{Feedback\} \newline Edit the user response such that the refinement doesn't have any errors mentioned in the feedback. Make the minimum number of changes when doing the refinement. Do not include a preamble.\\
        \midrule
        Multi-Agent Debate & \{Initial Prompt\} \newline Carefully review the following solutions from other agents as additional information, and provide your own answer and step-by-step reasoning to the question. \newline One agent's answer: \{\{``reasoning'': \{\}, ``answer'': \{\}\}\} \newline One agent's answer: \{\{``reasoning'': \{\}, ``answer'': \{\}\}\}\\
        \bottomrule
    \end{tabularx}
    \caption{Prompts for different subtasks and multi-agent debate.}
    \label{fig:prompts}
\end{table*}

\section{Human Evaluations}

\subsection{Critique Evaluation}\label{sec:human_eval_critique}

We sampled 50 examples and asked two authors to annotate the data using the same instructions provided to GPT, i.e., selecting from three choices. Annotators did not see the generated scores for any of the examples. We observed an inter-annotator agreement of 0.80 using macro-F1 and the average IAA between the GPT-predicted labels and our annotations is 0.61. This demonstrates that the GPT-based metric is an efficient and effective automatic evaluation method.

\subsection{Faithfulness Metric Correlations}\label{sec:human_eval_metric}
We conducted a blind, Likert scale human evaluation on 25 samples from MediaSum with MAMM-generated summaries, using the same prompt as for the GPT-based metric. Our annotated Likert scores achieved a Kendall correlation of 0.46 with the GPT Likert scores, which is comparable to the correlation reported in G-Eval \cite{liu-etal-2023-g}, a SOTA, Likert-based evaluation metric (0.43).

\section{Prompts}\label{sec:prompts}
We show the prompts for different pipelines in \autoref{fig:prompts}. We use the same 1-5 Likert prompt by \citet{dcr}, which contains a detailed rubric \cite{li2024leveraginglargelanguagemodels}.

\end{document}